\documentclass[11pt,a4paper]{article}
\usepackage[hyperref]{acl2018}
\usepackage{times}
\usepackage{latexsym}
\usepackage{amsmath}
\usepackage{url}
\usepackage{amssymb}
\usepackage{amsfonts}
\usepackage{graphicx}
\usepackage{tabularx}
\usepackage{multirow}

\aclfinalcopy 
\def\aclpaperid{3} 

\title{Convolutional neural networks for
chemical-disease relation extraction are improved with character-based word embeddings}

\author{Dat Quoc Nguyen \and Karin Verspoor  \\
School of Computing and Information Systems \\
The University of Melbourne, Australia \\
{\tt \{dqnguyen, karin.verspoor\}@unimelb.edu.au}
}

\date{}

\begin{document}
\maketitle
\begin{abstract}
We investigate the incorporation of character-based word representations into a standard CNN-based relation extraction model. 
We experiment with two common neural architectures, 
CNN and LSTM, 
to learn word vector representations from character embeddings. Through a task on the BioCreative-V CDR corpus, extracting relationships between chemicals and diseases, we show that models exploiting the character-based word representations improve on models that do not  use this information, obtaining state-of-the-art result relative to previous neural approaches.
\end{abstract}

\section{Introduction}\label{intro}

Relation extraction, the task of extracting  semantic relations between named entities mentioned in text, has become a key research topic in natural language processing (NLP) with a variety of practical applications \cite{bach2007}. Traditional approaches for relation extraction are feature-based and kernel-based supervised learning approaches which utilize various lexical and syntactic features as well as knowledge base resources; see the comprehensive survey of these traditional approaches in \newcite{surveyRE}. Recent research has shown that  neural network (NN) models for relation extraction obtain state-of-the-art performance. Two major neural architectures for the task include the convolutional neural networks, CNNs,  \cite{zeng-EtAl:2014:Coling,nguyen-grishman:2015,zeng-EtAl:2015:EMNLP,lin-EtAl:2016:P16-1,jiang-EtAl:2016:COLING1,zeng-EtAl:2017:EMNLP2017,huang-wang:2017:EMNLP2017} and long short-term memory networks, LSTMs \cite{miwa-bansal:2016:P16-1,zhang-zhang-fu:2017:EMNLP2017,katiyar-cardie:2017:Long,ammar-EtAl:2017:SemEval}. We also find  combinations of those two architectures \cite{Nguyen2016,Raj2017}. 

Relation extraction has attracted particular attention in the high-value biomedical domain. Scientific publications are the primary repository of biomedical knowledge, and given their increasing numbers, there is tremendous value in automating extraction of key discoveries~\cite{DEBRUIJN2002}. Here, we focus on the task of understanding relations between chemicals and diseases, which has applications in many areas of biomedical research and healthcare including toxicology studies, drug discovery and drug safety surveillance \cite{biocreativevcid}. 
The importance of chemical-induced disease (CID) relation extraction is also evident from the fact that chemicals, diseases and their relations are among the most searched topics by PubMed users~\cite{bap018}. In the CID relation extraction task formulation \cite{biocreativevcid,baw032}, CID relations are typically determined at document level, meaning that relations can be expressed across sentence boundaries; they can extend over distances of hundreds of word tokens.
As LSTM models can be difficult to apply to very long word sequences \cite{bradbury2016quasi}, 
CNN models may be better suited for this task. 

New domain-specific terms arise frequently in biomedical text data, requiring the capture of unknown words in practical relation extraction applications in this context. Recent research has  shown that character-based word embeddings enable capture of unknown words, helping to improve performance on many NLP tasks \cite{dossantos-gatti:2014:Coling,ma-hovy:2016:P16-1,lample-EtAl:2016:N16-1,plank-sogaard-goldberg:2016:P16-2,nguyen-dras-johnson:2017:K17-3}. This may be particularly relevant for terms such as gene or chemical names, which often have  identifiable morphological structure~\cite{krallinger2017information}.

We investigate the value of character-based word embeddings in a standard CNN model for relation extraction \cite{zeng-EtAl:2014:Coling,nguyen-grishman:2015}. 
To the best of our knowledge, there is no prior work addressing this. 

We experiment with two common neural architectures of CNN and LSTM for learning the character-based  embeddings, and evaluate the models on the benchmark BioCreative-V  CDR corpus  for  chemical-induced disease relation extraction \cite{baw068},  obtaining  state-of-the-art results.

\section{Our modeling approach}

This section describes our relation extraction models. They can be viewed as an  extension of the well-known CNN model for relation extraction \cite{nguyen-grishman:2015}, where we incorporate character-level representations of words.  

\begin{figure}[!ht]
\centering
\includegraphics[width=7.5cm]{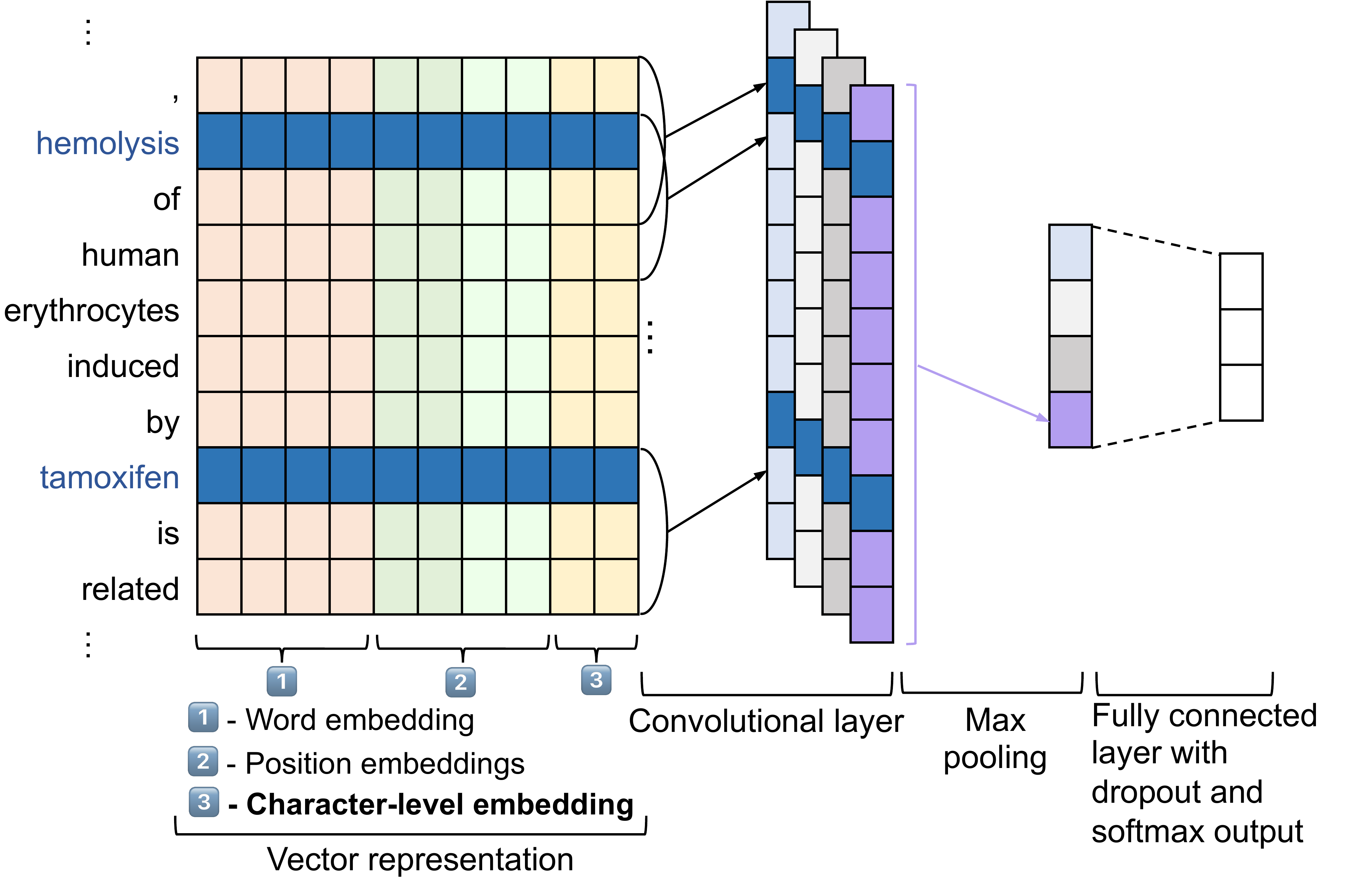} 
\caption{Our model architecture. Given the input relation mention marked with  two entities ``hemolysis'' and ``tamoxifen'', the convolutional layer uses the window size $k=3$ and the number of filters $m=4$.}
\label{fig:arch}
\end{figure}

Figure \ref{fig:arch}  presents our model architecture.  
Given an input fixed-length sequence (i.e.\ a \textit{relation mention}) of $n$ word tokens $w_1, w_2, w_3, ..., w_n$,\footnote{We set $n$ to be the length of the longest sequence and pad shorter sequences with a special ``PAD'' token.}    
marked with two entity mentions, the vector representation layer encodes each i$^{th}$ word  in the input relation mention by a real-valued vector representation $\boldsymbol{v}_i \in \mathbb{R}^{d}$. The convolutional layer takes the input matrix $\boldsymbol{S} = [\boldsymbol{v}_1, \boldsymbol{v}_2, ..., \boldsymbol{v}_n]^{\mathsf{T}}$ to extract high level features. These high level features are then fed into the max pooling layer to capture the most important features for generating a feature vector of the input relation mention. Finally, the feature vector is fed into a fully-connected neural network with softmax output to produce a probability distribution over relation types. For convenience, 
we detail the vector representation layer in  Section \ref{ssec:input} while the remaining layers appear in  Section \ref{ssec:cnnrl}. 

\subsection{CNN layers for relation extraction}\label{ssec:cnnrl}

\noindent\textbf{Convolutional layer:} This layer uses different filters to extract features from the input matrix $\boldsymbol{S} = [\boldsymbol{v}_1, \boldsymbol{v}_2, ..., \boldsymbol{v}_n]^{\mathsf{T}} \in \mathbb{R}^{n\times d}$ by performing convolution operations.  Given a window size $k$, a filter can be formalized as a weight matrix $\boldsymbol{F} = [\boldsymbol{f}_1, \boldsymbol{f}_2, ..., \boldsymbol{f}_k]^{\mathsf{T}}  \in \mathbb{R}^{k \times d}$.  For each filter $\boldsymbol{F}$, the convolution operation is performed to generate a feature map $\boldsymbol{x} = [\boldsymbol{x}_1, \boldsymbol{x}_2, ..., \boldsymbol{x}_{n-k+1}] \in \mathbb{R}^{n - k + 1}$:

\medskip
\centerline{$\boldsymbol{x}_j = g\big(\sum_{h=1}^{k} \boldsymbol{f}_h \boldsymbol{v}_{j+h-1} + b\big)$}
\medskip

\noindent where  $g(.)$ is some non-linear activation function and $b \in \mathbb{R}$ is a bias term.   

Assume that we use $m$ different weight matrix filters $\boldsymbol{F}^{(1)}$, $\boldsymbol{F}^{(2)}$, ..., $\boldsymbol{F}^{(m)} \in \mathbb{R}^{k \times d} $, the process above is then repeated $m$ times, resulting in $m$ feature maps $\boldsymbol{x}^{(1)}, \boldsymbol{x}^{(2)}, ..., \boldsymbol{x}^{(m)} \in \mathbb{R}^{n - k + 1}$. 

\noindent\textbf{Max pooling layer:} This layer aims to capture the most 
relevant features from each feature map $\boldsymbol{x}$ by applying the popular max-over-time pooling operation: 
$\hat{x} = \mathsf{max}\{\boldsymbol{x}\} = \mathsf{max}\{\boldsymbol{x}_1, \boldsymbol{x}_2, ..., \boldsymbol{x}_{n-k+1}\}$. From $m$  feature maps, the corresponding outputs are concatenated into a feature vector $\boldsymbol{z} = [\hat{x}^{(1)}, \hat{x}^{(2)}, ..., \hat{x}^{(m)}] \in \mathbb{R}^m$ to represent the input relation mention. 

\noindent\textbf{Softmax output:} The feature vector $\boldsymbol{z}$ is then fed into a fully connected NN followed by a  softmax layer for relation type classification. In addition, following \newcite{kim:2014:EMNLP2014},  for regularization  we apply dropout on $\boldsymbol{z}$ only during training. The softmax output procedure can be formalized as: 
 
\medskip
\centerline{$\boldsymbol{p} = \mathsf{softmax}\big( \mathbf{W}_1 (\boldsymbol{z} \ast \boldsymbol{r})+ \boldsymbol{b}_1\big)$}
\medskip

 \noindent where $\boldsymbol{p} \in \mathbb{R}^{t}$ is the final output of the network in which $t$ is the number of relation types, and $\mathbf{W}_1 \in \mathbb{R}^{t \times m}$ and $\boldsymbol{b}_1 \in \mathbb{R}^{t}$ are a transformation weight matrix and a bias vector, respectively. In addition, $\ast$ denotes an element-wise product and  $\boldsymbol{r} \in \mathbb{R}^{m}$ is a vector of independent Bernoulli random variables, each with probability  $\rho$ of being 0 \cite{JMLR:v15:srivastava14a}. 

\subsection{Input vector representation}\label{ssec:input}

This section presents the 
vector representation  $\boldsymbol{v}_i \in \mathbb{R}^{d}$  for each  i$^{th}$ word  token  in the  input relation mention $w_1, w_2, w_3, ..., w_n$. Let word tokens $w_{i_1}$ and $w_{i_2}$ be two entity mentions in the input.\footnote{If an entity spans over multiple tokens, we take only the last token in the entity into account \cite{nguyen-cho-grishman:2016:N16-1}.} 
We obtain $\boldsymbol{v}_i$  by concatenating word embeddings $\boldsymbol{e}_{w_i} \in \mathbb{R}^{d_1}$, position embeddings $\boldsymbol{e}^{(p1)}_{i - i_1}$ and  $\boldsymbol{e}^{(p2)}_{i - i_2}  \in \mathbb{R}^{d_2}$, and character-level embeddings $\boldsymbol{e}^{(\mathsf{c})}_{w_i}  \in \mathbb{R}^{d_3}$ (so, $d = d_1 + 2 \times d_2 + d_3$): 

\medskip
\centerline{$\boldsymbol{v}_i = \boldsymbol{e}_{w_i} \circ  \boldsymbol{e}^{(p1)}_{i - i_1} \circ \boldsymbol{e}^{(p2)}_{i - i_2} \circ \boldsymbol{e}^{(\mathsf{c})}_{w_i}   $}
\medskip


\noindent\textbf{Word embeddings:} Each word type $w$ in the training data is represented by a real-valued word embedding $\boldsymbol{e}_{w} \in \mathbb{R}^{d_1}$.  

\noindent\textbf{Position embeddings:} In relation extraction, we focus on assigning relation types to  entity pairs.  Words close to target entities are usually informative for identifying a relationship between them. Following \newcite{zeng-EtAl:2014:Coling}, to specify entity pairs, we use position embeddings $\boldsymbol{e}^{(p1)}_{i - i_1}$ and  $\boldsymbol{e}^{(p2)}_{i - i_2} \in \mathbb{R}^{d_2}$ to encode the relative distances $i-i_1$ and $i-i_2$ from each word  $w_i$ 
to entity mentions $w_{i_1}$ and $w_{i_2}$, respectively.  

\noindent\textbf{Character-level embeddings:} 
Given a word type $w$ consisting of $l$ characters $w = c_1c_2...c_l$ where each j$^{th}$ character in $w$  is represented by a character embedding  $\boldsymbol{c}_{j} \in \mathbb{R}^{d_4}$, we investigate two approaches for learning character-based word embedding $\boldsymbol{e}^{(\mathsf{c})}_{w} \in \mathbb{R}^{d_3}$ from  input $\boldsymbol{c}_{1:l}=[\boldsymbol{c}_1, \boldsymbol{c}_2, ..., \boldsymbol{c}_l]^{\mathsf{T}}$ as follows:

\smallskip
\noindent (1) Using \textbf{CNN} 
\cite{dossantos-gatti:2014:Coling,ma-hovy:2016:P16-1}: This CNN contains a convolutional layer to generate $d_3$ feature maps from the input $\boldsymbol{c}_{1:l}$, and  a max pooling layer to produce a final vector $\boldsymbol{e}^{(\mathsf{c})}_{w}$ from those feature maps  for  representing  the word $w$. 

\smallskip
\noindent (2) 
Using a sequence BiLSTM (\textbf{BiLSTM\textsubscript{seq}})
\cite{lample-EtAl:2016:N16-1}:  In  the BiLSTM\textsubscript{seq}, the input  is the sequence of  $l$ character embeddings $\boldsymbol{c}_{1:l}$, 
and the output is a concatenation of outputs of a forward LSTM (LSTM\textsubscript{f}) reading the input in its regular order and a reverse LSTM (LSTM\textsubscript{r}) reading the input in reverse:

\smallskip
\noindent\centerline{\small 
$\boldsymbol{e}^{(\mathsf{c})}_{w}  = \text{BiLSTM\textsubscript{seq}}(\boldsymbol{c}_{1:l}) \\ = \text{LSTM\textsubscript{f}}(\boldsymbol{c}_{1:l}) \circ \text{LSTM\textsubscript{r}}(\boldsymbol{c}_{l:1})$}


\subsection{Model training} 
The baseline CNN model for relation extraction \cite{nguyen-grishman:2015} is denoted here as \textbf{CNN}. The extensions incorporating CNN and BiLSTM character-based word  embeddings are \textbf{CNN+CNNchar} and \textbf{CNN+LSTMchar}, respectively.  The model parameters, including word, position, and character embeddings, weight matrices and biases, are learned during training to minimize the model negative log likelihood  (i.e.\ cross-entropy loss) with $L_2$ regularization. 

\section{Experiments}

\subsection{Experimental setup}

We evaluate our models using the BC5CDR corpus  \cite{baw068} which is the benchmark dataset for the chemical-induced disease (CID)  relation extraction task \cite{biocreativevcid,baw032}.\footnote{\url{http://www.biocreative.org/tasks/biocreative-v/track-3-cdr/}} The corpus consists of 1500 PubMed abstracts: 
500  for each of training, development and test.   
The training set is used to learn  model parameters, the development set to select optimal hyper-parameters, and the test set to report final results. We make use of  gold entity annotations in each case. 
 For evaluation results, we  measure the CID relation extraction performance with F1 score. More details of the dataset, evaluation protocol, and implementation are in the Appendix.

\subsection{Main results}

Table \ref{tab:results} compares the CID relation extraction results of our models to prior work. The first 11 rows report the performance of models that use the same experimental setup, without using additional training data or various features extracted from external knowledge base (KB) resources. The last 6 rows report results of models exploiting various kinds of features based on external relational KBs of chemicals and diseases, in which the last 4  SVM-based models are trained using both training  and  development sets.  

The models exploiting more training data and external KB features  obtained the best F1 scores. \newcite{Nagesh2016}  and \newcite{baw036} have shown that without KB features, their model performances (61.7\% and 67.2\%) are decreased by 5 and 11 points of F1 score, respectively.\footnote{\newcite{baw046} and \newcite{Peng2016} did not provide results without using the KB-based features. \newcite{baw036}  and \newcite{baw046} did not provide results in  using only the training set for learning models.}  Hence we find that external KB features are essential; we plan to extend our models to incorporate such KB features in future work.

\begin{table}[!ht]
\centering
\resizebox{7.75cm}{!}{
\setlength{\tabcolsep}{0.3em}
\begin{tabular}{llll}
\hline
\textbf{Model}  & \textbf{P} & \textbf{R} & \textbf{F1} \\
\hline
MaxEnt \cite{baw042} & 62.0 & 55.1 & 58.3 \\
Pattern rule-based \cite{baw039} & 59.3 & 62.3 & 60.8 \\
LSTM-based \cite{baw048} & {64.9} & 49.3 & 56.0\\
LSTM-based \& PP \cite{baw048} & 55.6 & 68.4 & 61.3 \\
CNN-based \cite{bax024} & 60.9 & 59.5 & 60.2\\
CNN-based \& PP \cite{bax024} & 55.7 & 68.1 & 61.3\\
BRAN \cite{akbc2017} & 55.6 & {70.8} & \textbf{62.1} \\
SVM+APG \cite{Panyam2018} & 53.2 & 69.7 & 60.3\\
\hline
CNN & 54.8 &{69.0} &	61.1 \\
CNN+CNNchar & {57.0} & 	68.6 & 	\textbf{62.3} \\
CNN+LSTMchar & 56.8 & 	68.8 & 62.2  \\
\hline
\hline
Linear+TK \cite{Nagesh2016} & 63.6 & 59.8 & 61.7  \\
SVM \cite{Peng2016} & 62.1 & 64.2 & 63.1 \\
SVM (+dev.) \cite{Peng2016} & 68.2 & 66.0 & 67.1 \\
SVM (+dev.+18K) \cite{Peng2016} & 71.1 & 72.6 & \textbf{71.8} \\
SVM (+dev.) \cite{baw036} & 65.8 & {68.6} & 67.2  \\
SVM  (+dev.) \cite{baw046} & {73.1} & 67.6 & {70.2} \\
\hline
\end{tabular}
}
\caption{Precision (P), Recall (R) and F1 scores (in \%). ``\& PP'' refers to  the  use of additional  post-processing heuristic rules. ``BRAN'' denotes bi-affine relation attention networks. ``SVM+APG'' denotes a model using SVM with All Path Graph kernel. ``Linear+TK'' denotes a  model combining linear and tree kernel classifiers. ``+dev.'' denotes the use of both training and development sets for learning models. Note that \newcite{Peng2016} also used an extra training corpus of 18K weakly-annotated PubMed articles.}
\label{tab:results}
\end{table}

In terms of models \textit{not} exploiting external data or KB features (i.e.\ the first 11 rows in Table \ref{tab:results}),  our  CNN+CNNchar and CNN+LSTMchar  obtain the highest F1 scores; with 1+\% absolute F1 improvements to the baseline CNN ($p$-value $<$ 0.05).\footnote{Improvements are significant with $p$-value $<$ 0.05 for a bootstrap significance test.} 
In addition, our models obtain  2+\% higher F1 score than the traditional feature-based models MaxEnt \cite{baw042} and SVM+APG \cite{Panyam2018}. We also achieve 2+\% higher F1 score than the LSTM- and CNN-based methods \cite{baw048,bax024} which exploit LSTM and CNN to learn relation mention representations from dependency tree-based paths.\footnote{\newcite{baw048} and \newcite{bax024} used the same post-processing heuristics to handle cases where models could not identify any CID relation between chemicals and diseases in an article, resulting in final F1 scores at 61.3\%.} 
Dependency trees have been actively used in traditional feature-based and  kernel-based methods for  relation extraction \cite{culotta-sorensen:2004:ACL,bunescu-mooney:2005:HLTEMNLP,GuoDong:2005:EVK:1219840.1219893,NIPS2005_2787,mintz-EtAl:2009:ACLIJCNLP} as well as in the biomedical domain \cite{btl616,Nagesh2016,Panyam2018,quirk-poon:2017:EACLlong}. Although we obtain better results, we believe dependency tree-based feature representations  still have strong potential value.  Note that to obtain dependency trees, previous work on CID relation extraction used the Stanford dependency parser \cite{chen-manning:2014:EMNLP2014}. However, this dependency parser was trained on the Penn Treebank (in the newswire domain) \cite{marcus1993ptb}; training on a domain-specific treebank such as CRAFT~\cite{bada2012craft}  
should help to improve results~\cite{verspoor2012corpus}.
 
We also achieve slightly better scores than the more complex model BRAN \cite{akbc2017}, the Biaffine Relation Attention Network, based on the Transformer self-attention model  \cite{NIPS2017_7181}. BRAN additionally uses byte pair encoding  \cite{Gage:1994:NAD:177910.177914} to construct  a vocabulary of subword units for tokenization. Using subword tokens to capture rare or unknown words has been demonstrated to be useful in machine translation  \cite{sennrich-haddow-birch:2016:P16-12} and likely captures similar information to character embeddings.  However, \newcite{akbc2017} do not provide comparative results using only original word tokens. Therefore, it is difficult to assess the usefulness specifically of using byte-pair encoded subword tokens in the CID relation extraction task, as compared to the impact of the full model architecture.  We also plan to explore the usefulness of  subword tokens in the baseline CNN  for future work, to enable comparison with the improvement when using the character-based word embeddings. 

It is worth noting that both CNN+CNNchar and CNN+LSTMchar return similar F1 scores, showing that in this case, using either CNN or BiLSTM to learn character-based word embeddings produces a similar improvement to the baseline.  There does not appear to be any reason to prefer one of these in our relation extraction application.

\section{Conclusion}
In this paper, we have explored the value of integrating character-based word representations into a baseline CNN model for relation extraction.  In particular, we investigate the use of two well-known neural architectures, CNN and LSTM, for learning character-based word representations. Experimental results on a benchmark chemical-disease relation extraction corpus show that the character-based  representations help improve the baseline to attain  state-of-the-art performance. Our models are suitable candidates to serve as future baselines for more complex models in the relation extraction task.

\paragraph{Acknowledgment:} This work was supported by the 
 ARC Discovery Project DP150101550.

\bibliography{Refs}
\bibliographystyle{acl_natbib}

\section*{Appendix}

\paragraph{Dataset and evaluation protocol:} 
We evaluate our models using the BC5CDR corpus  \cite{baw068}, which is the benchmark dataset for the BioCreative-V  shared task on chemical-induced disease (CID)  relation extraction \cite{biocreativevcid,baw032}.\footnote{\url{http://www.biocreative.org/tasks/biocreative-v/track-3-cdr/}} The BC5CDR corpus consists of 1500 PubMed abstracts: 
500 each for training, development and test set.   
In all articles, chemical and disease entities were manually annotated using the Medical Subject Headings (MeSH) concept identifiers \cite{mesh}. 

\begin{figure}[t]
\centering
\includegraphics[width=7.5cm]{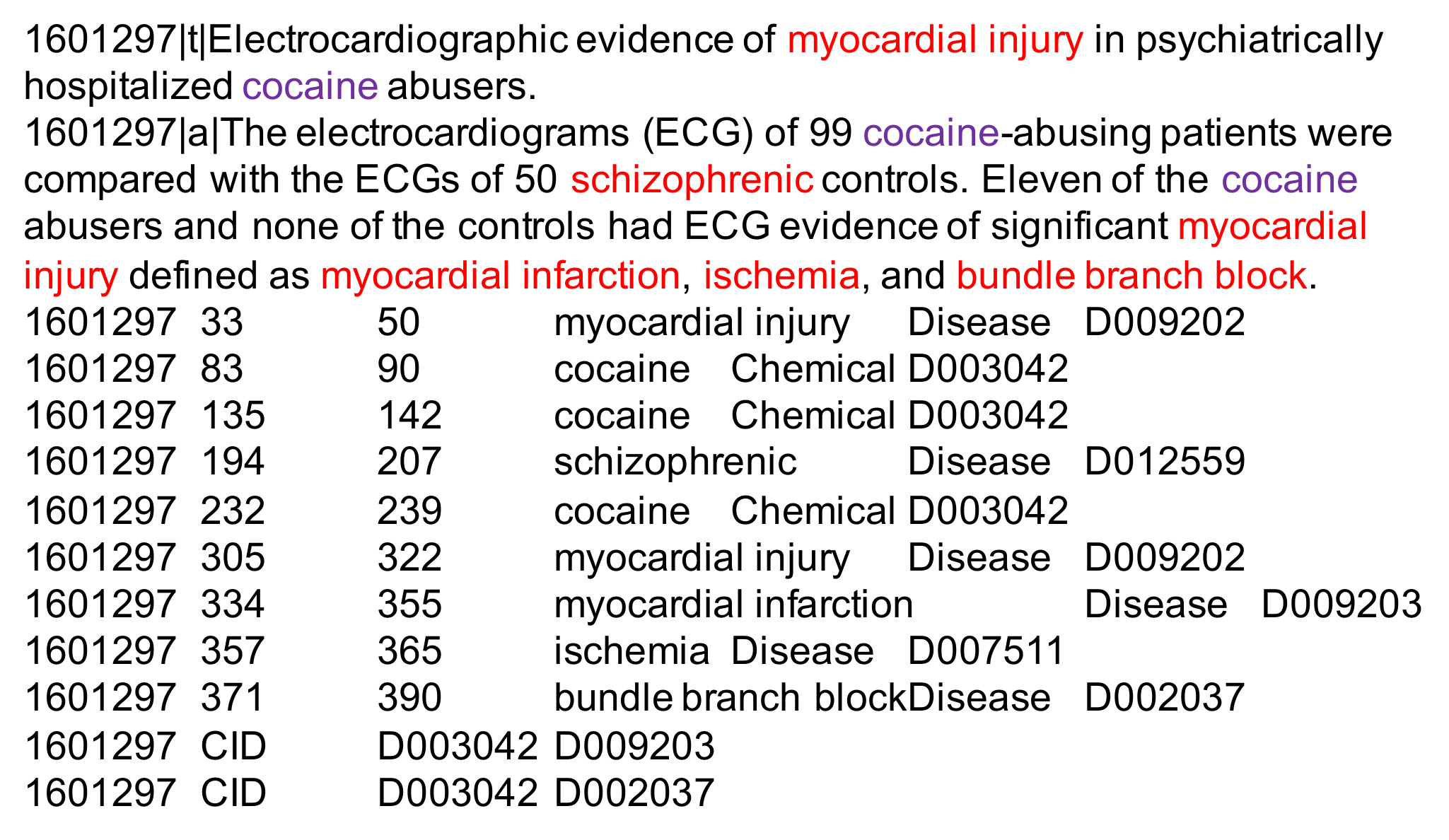} 
\caption{A part of an annotated PubMed article.}
\label{fig:example}
\end{figure}

CID relations were manually annotated for each relevant pair of chemical and disease concept identifiers at the \textit{document level} rather than for each pair of entity mentions (i.e.\ the relation annotations are not tied to specific mention annotations).  Figure \ref{fig:example} shows  examples of CID relations. 
We follow \newcite{baw042} (see relation instance construction and hypernym filtering sections) and \newcite{bax024} to transfer these annotations to \textit{mention level} relation annotations.  


In the evaluation phase, mention-level classification decisions must be transferred to the document level. Following \newcite{baw042}, \newcite{7822658} and \newcite{bax024}, these are derived from either
(i) a pair of entity mentions that has been positively classified to form a CID relation based on the document
or
(ii) a pair of entity mentions that co-occurs in the document, and that has been annotated as having a CID relation in a document in the training set.

In an article, a pair of chemical and disease concept identifiers may have multiple entity mention pairs, expressed in different relation mentions. 

The longest relation mention has about 400 word tokens; the longest word has 37 characters. 

We use the training set to learn  model parameters, the development set to select optimal hyper-parameters, and the test to report final results using gold entity annotations. 
 For evaluation results, we  measure the CID relation extraction performance using F1 score.
 
\paragraph{Implementation details:} We implement CNN, CNN+CNNchar, CNN+LSTMchar using Keras \cite{chollet2015keras} with a TensorFlow backend \cite{45381}, and  use a fixed random seed.
For both CNN+CNNchar and CNN+LSTMchar, character embeddings are randomly  initialized with 25 dimensions, i.e.\ $d_4 = 25$. 
For CNNchar, 
the window size is 5 and the number of filters at 50, resulting in $d_3 = 50$. For LSTMchar, we  set the number of LSTM units at 25,  also resulting in $d_3 = 50$. 

For all three models, position embeddings are randomly initialized with 50 dimensions, i.e.\ $d_2 = 50$.  
Word embeddings are initialized by using 200-dimensional pre-trained word vectors from \newcite{chiu-EtAl:2016:BioNLP16}, i.e.\ $d_1 = 200$; and word types (including a special ``UNK'' word token representing unknown words), which are not in the embedding list, are initialized randomly. Following \newcite{TACL885}, the  ``UNK'' word embedding is learned during training by replacing each word token $w$ appearing $n_w$ times in the training set  with ``UNK'' with probability $\mathsf{p}_{unk}(w) = \frac{0.25}{0.25 + n_w}$ (this  procedure only involves the word embedding part in the input vector representation layer). 
 We use ReLU for the activation function $g$,  and fix the window size $k$ at 5 and the $L_2$ regularization value at 0.001.

We train the models with Stochastic gradient descent  using  Nadam \cite{Dozat2015IncorporatingNM}.  
For training, we run for 50 epochs. We perform a grid search to select the optimal hyper-parameters by monitoring the F1 score after each training epoch on the development set. Here, we select the  initial Nadam learning rate  $\lambda \in \{\text{5e-06}, \text{1e-05}, \text{5e-05}, \text{1e-04}, \text{5e-04}\}$,  the number of filters $m \in \{100, 200, 300, 400, 500\}$ and the dropout probability $\rho \in \{0.25, 0.5\}$.  We choose the model with highest  F1  on the development set, which is then applied to the test set for the evaluation phase. 



\end{document}